\documentclass[letterpaper, 10 pt, conference]{ieeeconf}  

\IEEEoverridecommandlockouts   

\title{\LARGE \bf
Triplane Grasping: Efficient 6-DoF Grasping with Single RGB Images
}

\author{Yiming Li$^{1}$, Hanchi Ren$^{1}$, Yue Yang$^{1}$, Jingjing Deng$^{2}$ and Xianghua Xie$^{1}$*
\thanks{$^{1}$Yiming Li, Hanchi Ren, Yue Yang, and Xianghua Xie are with the Computer Science Department, Faculty of Science and Engineering, Swansea University, UK}
\thanks{$^{2}$Jingjing Deng is with Department of Computer Science, Durham University, UK}%
\thanks{$^{*}$Corresponding Author: Xianghua Xie ({\tt\small x.xie@swansea.ac.uk})}
\thanks{This work is supported by the EPSRC National Edge AI Hub (EP/Y007697/1).}
}

\usepackage{cite}
\usepackage{color}
\usepackage{amsmath}
\usepackage{multirow}
\usepackage{booktabs}
\usepackage[table]{xcolor}
\usepackage{graphicx}

\definecolor{lightorange}{RGB}{255, 230, 204} 
\definecolor{darkorange}{RGB}{255, 178, 102}   

\begin{document}

\maketitle
\thispagestyle{empty}
\pagestyle{empty}

\begin{abstract}
Reliable object grasping is one of the fundamental tasks in robotics. However, determining grasping pose based on single-image input has long been a challenge due to limited visual information and the complexity of real-world objects. In this paper, we propose Triplane Grasping, a fast grasping decision-making method that relies solely on a single RGB-only image as input. Triplane Grasping creates a hybrid Triplane-Gaussian 3D representation through a point decoder and a triplane decoder, which produce an efficient and high-quality reconstruction of the object to be grasped to meet real-time grasping requirements. We propose to use an end-to-end network to generate 6-DoF parallel-jaw grasp distributions directly from 3D points in the point cloud as potential grasp contacts and anchor the grasp pose in the observed data. Experiments on the OmniObject3D and GraspNet-1Billion datasets demonstrate that our method achieves rapid modeling and grasping pose decision-making for daily objects, and strong generalization capability.

\end{abstract}

\section{INTRODUCTION}

\subsection{From image to 3D}


The task of recovering a 3D model of an object from a single image is inherently ambiguous because it lacks sufficient information to make restoring a high-quality, well-textured representation a challenge. Recently diffusion models \cite{c1,c2}, which have received a lot of attention, have become a popular approach to address such challenges due to their excellent 2D generation performance. By leveraging the novel view images generated by diffusion models, the single-image object reconstruction task can be transformed into a multi-view reconstruction task to improve the overall quality of the resulting 3D model \cite{c3,c4,c5}. However, although most of the methods in this category have multi-view attention mechanisms or multi-view correlation strategies to improve view consistency, they still essentially extract information from a single image for reconstruction and lack sufficient 3D structural constraints. Such geometrically unstable reconstruction is unacceptable for tasks based on 3D point clouds for grasping.

To enhance the structural reconstruction performance, alternative methods utilize prior shape knowledge tailored to specific categories for deriving 3D forms from single images \cite{c6,c7}. Such methods alleviate the problem of missing 3D structural constraints by learning strong geometric priors for different classes of objects. Meanwhile, based on different needs such as fast generation or high-quality reconstruction, many different 3D representations are used to match different tasks. Categorically, they include explicit (e.g. point cloud \cite{c8,c9,c10}, mesh \cite{c11,c12}, voxel \cite{c13,c14}, 3D Gaussian \cite{c15}), implicit (notably Neural Radiance Fields or NeRF \cite{c16}) and hybrid (Triplane \cite{c17}) representations. NeRF has recently gained a lot of attention for its ability in detailed and textural reconstruction.
However, due to the dense sampling and multiple network query characteristic of volume rendering, as well as the significant computational effort required for the numerical approximation process of the volume rendering integrals, pipelines based on NeRF representations tend to be significantly more time-consuming than other representations, both in terms of training and rendering time. There have been various follow-up works proposing optimization strategies for different stages, such as hash encoding \cite{c18}, sparse voxel fields \cite{c19} and replacing NeRF network itself with multiple tiny MLPS \cite{c20}, etc. 
Some of the methods have demonstrated significant speedups; however, due to the implicit nature of NeRF, re-training is required when facing a new scene. 
This makes them unsuitable for real-time object grasping.


The Triplane representation has garnered significant attention recently due to its exceptional performance and hybrid characteristics \cite{c21}. This approach utilizes only three mutually perpendicular 2D planes to encode 3D information, substantially reducing computational complexity. Moreover, the distribution of 3D feature information across these three 2D planes facilitates rapid querying of features at arbitrary 3D positions, enabling efficient ray tracing and rendering. However, Triplane employs volume rendering similar to NeRF, which still presents challenges in terms of memory consumption and real-time reconstruction and rendering.

Another notable approach to 3D reconstruction is Gaussian Splatting \cite{c22}. This explicit 3D representation method diverges from volume rendering techniques that require dense sampling and weighted approximations to obtain novel view pixel RGB values for 3D objects at specific angles. Instead, it employs a multitude of parameterized 3D Gaussian ellipsoids to constitute objects or scenes. These ellipsoids are controlled by various parameters, including position, rotation, and spherical harmonics (SH) coefficients, and may overlap or contact each other. The rapid rendering capability of 3D Gaussian Splatting is attributed to its explicit nature. This point-based representation method only requires the camera angle to be specified, allowing for the collection of all visible 3D Gaussian ellipsoids within a specific resolution range along the incident direction. By weighting these ellipsoids based on their transparency, color, and other attributes, novel view images can be efficiently generated.

Recently, there are significant interests in leveraging large-scale 3D datasets to learn robust 3D shape or texture priors, e.g. \cite{c24,c25,c26}.
For instance, capitalizing on the high-capacity characteristics of transformer architectures \cite{c27}, LRM \cite{c28} introduced a fully transformer-based framework, which enables the extraction of image features encoded via DINO \cite{c29} as triplane tokens, and subsequently decoded into NeRF representations. Similarly, Instant3D \cite{c30} employs transformer to reconstruct NeRF representations from four-view images.
Such 3D shape prior learning approaches, along with other category-specific methods (e.g. 3D templates \cite{c31}) has shown promising performance on generalization ability within the category range, but often fails in providing quality reconstruction results when facing unseen categories or styles.

\subsection{Robotic arm grasping}

Robotic grasping research can be broadly categorized into two main approaches: model-based and model-free grasping. In model-based approaches, the 3D model or type of the object is known \emph{a priori}. 
Typically, model-based methods first identify key points on the target object and then select an appropriate transformation from a set of predefined grasping actions to execute the grasp \cite{c39,c40}. The primary advantage of this approach lies in its ability to circumvent the need for inferring potential grasping points and the physical properties of the target object. However, model-based methods heavily rely on predefined object models, which often results in limited generalization capabilities, particularly when confronted with novel or previously unseen objects.
Consequently, these strategies are typically best suited for category-specific task environments, where precisely maintained object model libraries and high-precision sensors facilitate relatively fixed grasping or object manipulation tasks. The limitations of model-based approaches highlight the need for more adaptive and generalizable grasping strategies, particularly in dynamic and unpredictable environments. 

Model-free grasping offers a more flexible and adaptive paradigm. 
However, the pursuit of suitable universal object representations often necessitates training on large-scale, data-driven grasping constraints. This requirement stems from the vast search space encompassing all possible 4-DoF or 6-DoF grasping poses for arbitrary objects. Consequently, developing models capable of learning high-success-rate, reasonable grasping poses remains a significant challenge in the field. Furthermore, the geometric structure and performance characteristics of the gripper itself introduce additional spatial constraints.
As a result, it is often necessary to generate a series of possible grasping poses for an object rather than a single solution, taking into account the physical constraints of the robotic arm.

Due to the data-dependent nature of model-free approaches, the majority of works rely on dense input data. This often involves multi-view 3D reconstruction in multi-camera environments to aid in 3D point cloud generation of objects or scenes, or the use of video frames surrounding the target point or object to dynamically construct the scene and detect possible object poses. Currently, only a small number of studies focus on single-image input grasping scenarios, and some of these implementations still depend on depth maps provided by sensors such as ToF and LiDAR. This phenomenon primarily stems from the fact that robust and complete restoration of an object's 3D structure significantly impacts the success rate of subsequent object pose detection and grasp action generation. 

\subsection{6-DoF grasping}
Six degree of freedom (6-DoF) grasping has become increasingly crucial in the field of robotic manipulation, offering a powerful tool for addressing grasping challenges in complex scenarios. In contrast to traditional planar grasping, 6-DoF grasping enables robots to precisely control the position and orientation of their end effectors in three-dimensional space, significantly enhancing grasping flexibility and adaptability. This high degree of freedom allows robots to effectively handle various complex object shapes and poses, even in confined or cluttered spaces.

Common approaches to implementing 6-DoF grasping can be broadly categorized into depth-based and RGB-based methods. For instance, the ``6-DoF Contrastive Grasp Proposal Network" utilizes a single depth image as input, employing a proposed rotated region proposal network in conjunction with a collision detection module to infer 6-DoF grasping actions for parallel-jaw grippers. However, such works are often limited to simple objects or scene setups, demonstrating significant limitations when faced with more complex operations and environments. Another evident drawback is their strong reliance on depth sensors, and their potential failure when dealing with soft or transparent objects due to the lack of object image information.

In pursuit of more user-friendly and versatile frameworks, some implementations focus on end-to-end training strategies using reinforcement learning. These approaches take images as input and directly output grasping poses \cite{c42,c43}. The primary advantage of such methods lies in their ability to be rapidly deployed on different devices with minimal or no retraining, allowing for immediate commencement of grasping operations. These approaches generally demonstrate good performance in simple planar and non-complex scenarios, showcasing excellent capabilities under specific scenes or requirements, but may face challenges in more complex or dynamic environments. 

To meet operational requirements in complex scenarios, related works such as VoteGrasp\cite{c48} utilize contextual information, encoding the dependencies of recognized objects in the scene as features, and combining them with given scene point cloud information to generate collision-free 6-DoF point cloud grasping poses. However, these works often lack consideration of object geometric shape priors, unable to further refine grasp predictions. Some works have a similar pipeline to our proposed method, such as TransSC\cite{c49}, which first uses a transformer-based encoder and a manifold-based decoder to transform segmented partial point clouds into more complete and detailed object representations, and then integrates the reconstructed fine object representations into the grasp evaluation network to generate final grasp proposals.

The works most similar to our method is MonoGraspNet\cite{c50} and CenterGrasp\cite{c51}. MonoGraspNet uses a keypoint heatmap and a normal map to recover the 6-DoF grasping poses using a new parameterized representation. CenterGrasp, on the other hand, uses a traditional encoder-decoder structure, following PanopticFPN\cite{c52} in the encoder part, which includes a heatmap head, a pose head, and a latent code head. In the decoder part, it follows the DeepSDF\cite{c53} to obtain shape and grasp pose from a coordinate in space and the output of the latent code head. Both of these works achieve grasp generation from single image input, but they either still need to rely on depth map information for more detailed grasp optimization or have not achieved real-time grasp decision-making requirements.

\section{Proposed Method}

\begin{figure*}[ht]
\centering
\includegraphics[width=1\textwidth]{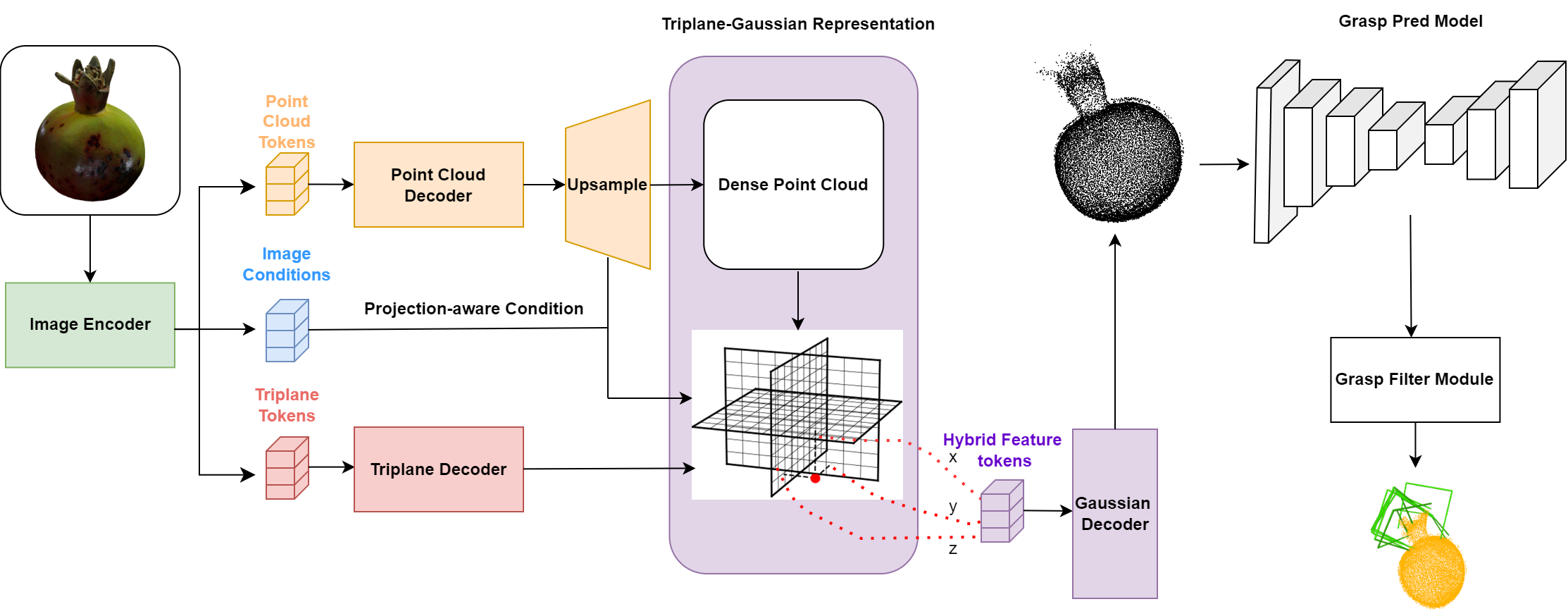}
\caption{{The pipeline of Triplane Grasping. The input RGB image was first encoder by the pre-trained DINOv2\cite{c54} image encoder which output point cloud tokens, image conditioning tokens and triplane representation tokens. After the decoding process of point cloud decoder along with the triplane decoder, a hybrid Triplane-Gaussian representation is obtained for the further query request from Gaussian Decoder. The generated point cloud will then feed into the grasp prediction model with a contact filtering module to refine the final grasp decision.}}
\label{fig:arch}
\end{figure*}

In this paper, we introduce Triplane Grasping, a novel framework capable of rapidly generating grasping poses from a single RGB image. Triplane Grasping achieves efficiency and robustness through its unique two-stage design: first, a 3D object reconstruction module efficiently and accurately reconstructs the object's 3D point cloud representation from a single RGB image; subsequently, a pose detection module predicts the 6-Dof grasping base on the reconstructed point cloud. 
Our main contribution lies in the simultaneous resolution of two major challenges faced by existing methods: the trade-off between speed and quality, and limitations in generalization capability. Unlike previous methods that achieve high-quality reconstruction and grasping at the cost of speed, Triplane Grasping significantly enhances processing speed while maintaining performance. Compared to other efficient grasping methods that compromise on reconstruction quality or grasping success rates, Triplane Grasping maintains produces high quality grasping decisions. Most importantly, Triplane Grasping demonstrates excellent generalization capabilities, i.e. rapidly generating high-quality 3D point clouds and producing robust grasping poses in various unseen scenarios. This feature makes it particularly suitable for real-world applications requiring quick adaptation to new environments, offering a possible direction in the field of single-image-driven robotic grasping in achieving efficient and adaptable robotic manipulation in diverse settings. 

Figure 1 shows the overall two-stage pipeline of our method. In the first 3d reconstruction stage, we conduct a quick point cloud regression via a transformer decoder using the extracted latent features token from DINOv2 \cite{c54} as the input. A Triplane decoder, also based on a Transformer architecture, generates an implicit volumetric feature field in the form of three axis-aligned 2D feature planes, conditioned on both the image features and the previously reconstructed geometry via geometry-aware encoding. Finally, for each point in the point cloud, its corresponding features are extracted from the triplane and combined with local image features through projection-aware conditioning, and then decoded via a lightweight MLP to estimate the parameters of anisotropic 3D Gaussians, including opacity, spatial offset, scale, rotation, and spherical harmonics coefficients. The final reconstructed point clouds are rendered using a fast and differentiable Gaussian Splatting, thus achieving efficient single RGB 3D point cloud reconstruction. 

To facilitate downstream robotic interaction, we further leverage the power of learned grasp generation frameworks to predict robust and diverse 6-DoF grasp poses directly from the reconstructed point cloud. Specifically, the grasping model treats each point in the scene as a potential contact candidate and infers grasp configurations by associating semantic and geometric cues in local 3D neighborhoods. By anchoring grasp predictions to the complete reconstructed point cloud geometry, the method enables grasp planning even for parts of the object that are not visible from the original input view, overcoming limitations typically observed in approaches that rely solely on partial, view-dependent observations. Moreover, the grasp generation component seamlessly integrates into our pipeline, providing a lightweight and computationally efficient solution that aligns well with the overall goal of fast and scalable inference. The technical details will be introduced in the following.

\subsection{3D Reconstruction }



\subsubsection{Hybrid Triplane-Gaussian Representation}

We first use a pre-trained DINOv2 \cite{c54} to obtain patch-wise feature tokens from the image. Additionally, camera information is encoded as camera features through adaptive layer normalization and incorporated into DINOv2.
These features, fused with camera pose data, are then passed to a transformer-based point cloud decoder, which generates a coarse point cloud with 2048 points. A two-step Snowflake point deconvolution (SPD) \cite{c56} is applied for densify the point cloud further to 16384 points. 

To remain reconstruction efficient as well as quality, following \cite{c23}, we adopt the hybrid Triplane-Gaussian Representation, where explicit point cloud and implicit Triplane Feature Field are jointly leveraged. The Triplane is a compact 3D representation that implicitly represents 3D structures by means of three axis-aligned orthogonal planes. Such implicit representation of feature encoding by only three orthogonal planes provides efficient feature querying capabilities. Following such design, the previous dense point cloud and image latent features will then pass into a transformer-based triplane decoder to produce triplane tokens, encoding 3D information using three orthogonal planes. These triplane features are compact yet informative, enabling efficient querying of spatial data.


Specifically, the explicit part of the hybrid representation is represented by the point cloud $P \in {R}^{N \times 3}$ to provide geometric information about the object, and Triplane $T \in {R}^{3 \times C \times H \times W}$ then implicitly encodes the eigenfields that can decode the 3D. For any point in space $x\in {R}^{3}$, its corresponding Triplane feature vector can be obtained by trilinear interpolation:

\begin{equation}
\mathbf{f_t} = \text{interp}(T_{xy}, p_{xy}) \oplus \text{interp}(T_{xz}, p_{xz}) \oplus \text{interp}(T_{yz}, p_{yz})
\end{equation}
where $T_{xy}$,$T_{xz}$ and $T_{yz}$ represent for three axis-aligned feature planes from Triplane Decoder, $ \emph{interp}$ as the trilinear interpolation and $\oplus$ as feature concatenation.

After obtaining the triplane feature vector, for any point coordinate $x \in {R}^{3}$ in the point cloud, an MLP is deployed to decode the query feature $f$ from triplane, which represents the 3DGS attributes of the coordinate:

\begin{equation}
(\mathbf{p}, \alpha, \mathbf{s}, \mathbf{r}, \mathbf{sh}) = \varphi_g(\mathbf{x}, \mathbf{f})
\end{equation}
where $\mathbf{p}, \mathbf{s}, \mathbf{r},\alpha, \mathbf{sh}$ denote the position (xyz coordinates), scale, rotation, opacity, and spherical harmonic (SH) coefficient, respectively.

\subsubsection{Point Cloud Decoder \& Triplane Decoder}

Both Point Cloud Decoder and Triplane Decoder are purely transformer-based architecture. Each transformer block contains a self-attention layer, a cross-attention layer as well as a feed-forward layer. For Point Cloud Decoder, which is aimed for constructing the coarse shape of the object, we follow similar design as Point-E \cite{c9} but in a feed-forward fashion, that decodes a point cloud through image conditions from point cloud tokens. The number of points to decode from the coarse point cloud representation is set to 2048 in an effort to boost memory efficiency.

The Triplane Decoder generates an implicit feature field by combining information from both the image and the initial point cloud. This feature field is then used by the 3D Gaussian Decoder to query and decode Gaussian attributes such as opacity and anisotropic covariance. To further refine geometry, the upsampled point cloud is re-encoded into the decoder using its positional embeddings, ensuring a closer alignment between the implicit triplane features and the explicit point cloud geometry.

To effectively leverage both 3D structural information and 2D perceptual cues, we build our training objective upon geometry-based losses while incorporating additional image-level supervision. Specifically, the overall loss function is composed of two parts: a geometry loss $\mathcal{L}_{\text{geo}}$ and a rendering loss $\mathcal{L}_{\text{render}}$:

\begin{equation}
\mathcal{L} = \mathcal{L}_{\text{geo}} + \mathcal{L}_{\text{render}}
\end{equation}

The geometry loss is formulated as:
\begin{equation}
\mathcal{L}_{\text{geo}} = \lambda_c \cdot \mathcal{L}_{\text{CD}} + \lambda_e \cdot \mathcal{L}_{\text{EMD}}
\end{equation}
where $\mathcal{L}_{\text{CD}}$ and $\mathcal{L}_{\text{EMD}}$ denote the Chamfer Distance and Earth Mover’s Distance between the predicted and ground-truth point clouds, respectively. These terms enforce accurate spatial alignment in 3D space. In our implementation, we set $\lambda_c=10$ and $\lambda_e=10$.

To further enhance the expressiveness of the learned representation, we integrate a 2D rendering supervision loss based on differentiable rendering. The rendered images from predicted 3D structures are compared against ground-truth views using perceptual and structural consistency metrics:
\begin{equation}
\mathcal{L}_{\text{render}} = \frac{1}{N} \sum_{i=1}^{N} \left( \mathcal{L}_{\text{MSE}}^{(i)} + \lambda_s \cdot \mathcal{L}_{\text{SSIM}}^{(i)} + \lambda_l \cdot \mathcal{L}_{\text{LPIPS}}^{(i)} \right)
\end{equation}
where $\mathcal{L}_{\text{MSE}}^{(i)}$, $\mathcal{L}_{\text{SSIM}}^{(i)}$, and $\mathcal{L}_{\text{LPIPS}}^{(i)}$ are computed between the $i$-th rendered view and its corresponding ground-truth image. We set $\lambda_s=1$ and $\lambda_l=2$ in all experiments.

\subsection{Efficient Grasping}

To better align with the goal of efficient grasping generation, We apply Contact-GraspNet \cite{c44} to generate 6-DoF grasp poses based on the reconstructed complete object point clouds. Specifically, Traditional 6-DoF grasp learning problems often require regression in SE(3) space, which is high-dimensional and discontinuous, making it difficult to optimize. We only consider the 3D points on the object surface as potential grasp contact points, significantly simplifying the process of learning grasp poses in the first place.  By reducing the learning problem to contact point classification and grasp rotation estimation, the dimensionality of the grasp representation is compressed to 4-DoF, which greatly enhances learning efficiency and grasp pose accuracy.

Mathematically, given a contact point \(c \in \mathrm{R}^3\), the grasp pose \(g \in SE(3)\) is defined by the rotation matrix \(R_g \in \mathrm{R}^{3 \times 3}\) and the translation vector \(t_g \in \mathrm{R}^3\). The rotation matrix is constructed from two orthogonal unit vectors, the grasp baseline direction \(b \in \mathrm{R}^3\) and the grasp approach direction \(a \in \mathrm{R}^3\), where \( ||a|| = ||b|| = 1 \). The parameterized form of the grasp pose \(g\) is as follows:

\begin{equation}
t_g = c + \frac{w}{2} \mathbf{b} + d \mathbf{a}
\end{equation}

\begin{equation}
R_g = \begin{bmatrix} 
\vert & \vert & \vert \\
\mathbf{b} & \mathbf{a} \times \mathbf{b} & \mathbf{a} \\
\vert & \vert & \vert
\end{bmatrix},
\end{equation}

where \(w\) is the grasp width, and \(d\) is the fixed distance from the gripper baseline to the base frame. This grasp representation not only reduces the degrees of freedom for the predicted poses but also leverages the geometry of the point cloud, ensuring that the predicted grasp poses align with the contact points on the object's surface. Furthermore, the previous fast 3D reconstruction stage replaced the requirement of the original RGB-D image as input, further boost the speed of the entire pipeline. 

\begin{figure}[ht]
\centering
\includegraphics[width=\columnwidth]{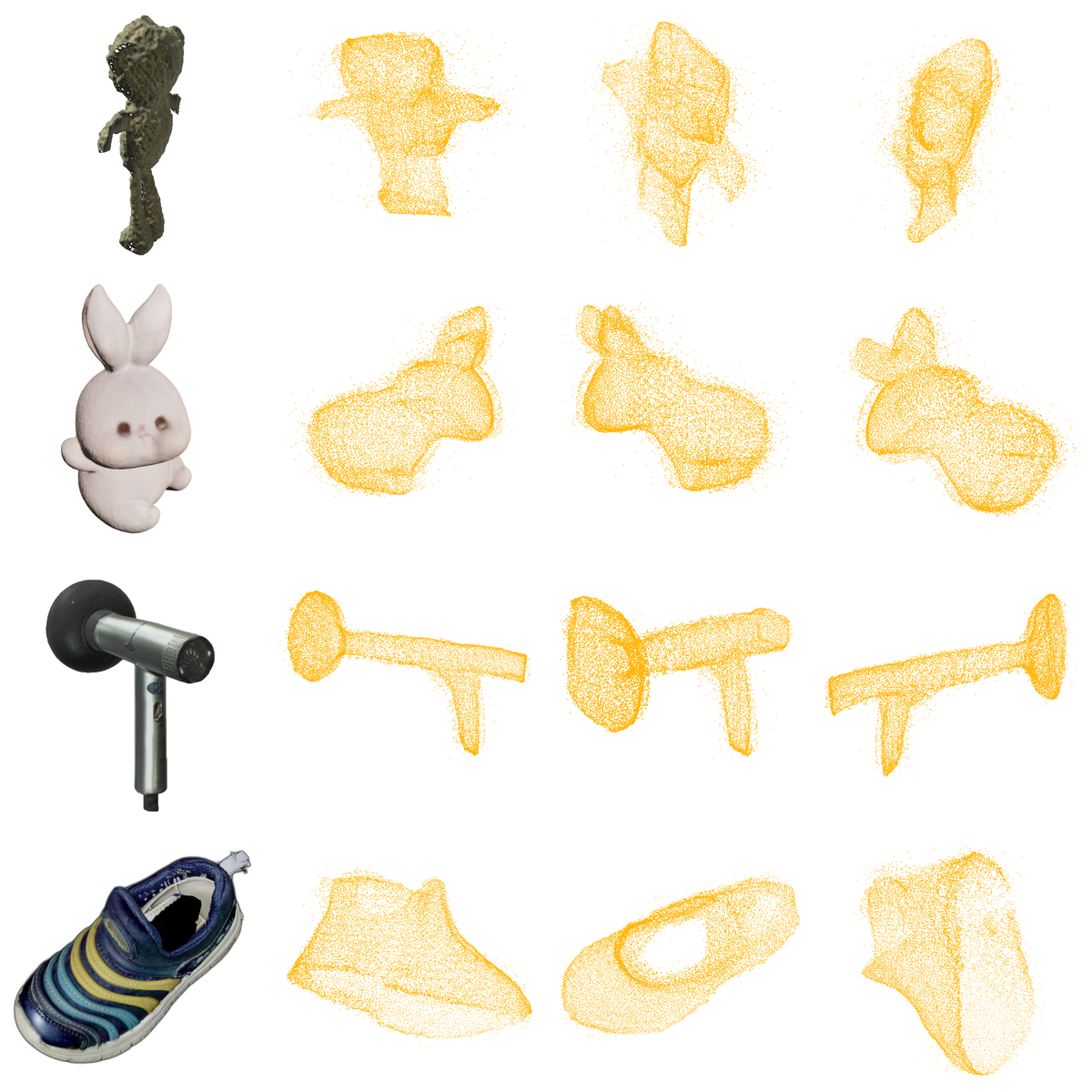}
\caption{Visualization of the reconstructed point cloud from single images.}
\label{fig:rec_result}
\end{figure}

In addition to the grasp pose generation, our method employs grasp contact filtering to ensure that the generated grasps are associated with the target object. After generating grasps from the full scene or local region point cloud, the grasp contact filtering step is used to associate the predicted grasps with specific object segments. This process involves filtering out grasps whose contact points do not belong to the desired target object. By filtering based on the contact points of each predicted grasp, the method ensures that only those grasps whose contact points lie on the target object are retained. This effectively eliminates grasps that could cause collisions with surrounding objects or unintended interactions with the environment, thus improving the overall grasp success rate.

\begin{figure}[ht]
\centering
\includegraphics[width=\columnwidth]{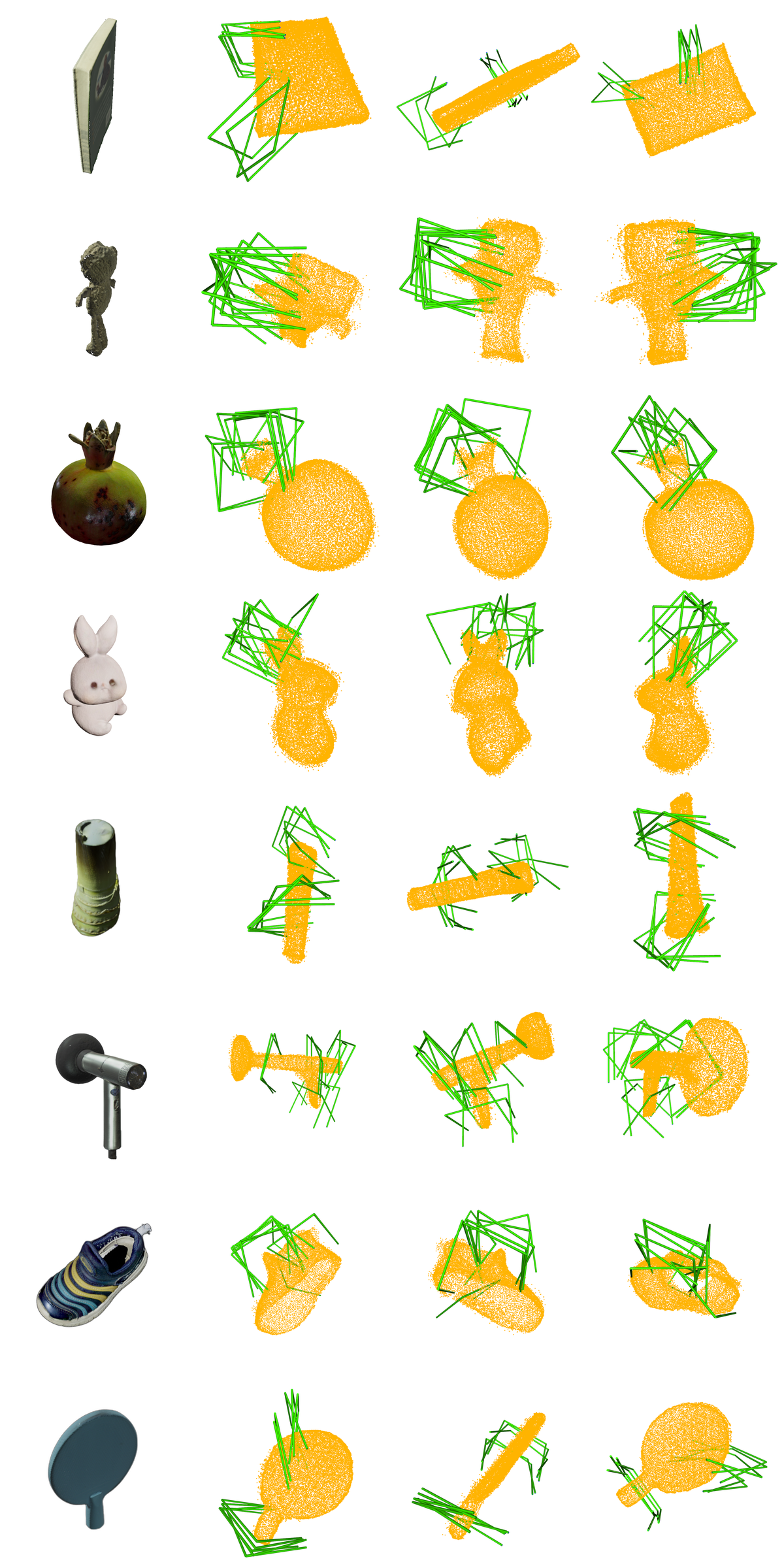}
\caption{Visualization of Triplane Grasping.}
\label{fig:vis}
\end{figure}

\section{Experiment}

\subsection{Dataset} 
We use the OmniObject3D dataset~\cite{c55} as the training set of our method, to further evaluate the generalization capability and the effectiveness of our method for grasp prediction, we conduct additional experiments on the GraspNet-1Billion dataset~\cite{c57}. OmniObject3D is a large-scale dataset contains 5,910 objects distributed across 216 classes of common desktop objects. The dataset contains complex objects, in terms of geometry, topology and appearance. The GraspNet-1Billion dataset~\cite{c57} contains 190 scenes, each captured from 256 different viewpoints with corresponding RGB and depth images. It covers 88 distinct object categories. Among the 190 scenes, 100 are split as the training set and the remaining 90 are reserved as the testing set. The testing set is further divided into three subsets: seen, similar, and novel, which correspond to test scenes containing objects that are either present in the training set, similar to those in the training set, or completely unseen during training, respectively.

\subsection{3D Reconstruction}

The accuracy of 3D reconstruction is highly correlated with the performance of grasp prediction based on it. Therefore, We first evaluate the performance of our method in reconstructing unseen objects. To this end, we split the objects in the OmniObject3D dataset into 80\% for training and 20\% for testing. Table~\ref{tab:performance_comparison} presents a comparative evaluation of different single-view 3D point cloud reconstruction methods based on a single RGB image.
The evaluation metrics we present include Chamfer Distance (CD) and F-Score (FS), where F-Score is calculated using a threshold of 0.02. For fair evaluation across methods with differing scale capabilities, both predicted and ground-truth aare normalized prior to computing Chamfer Distance and F-Score. Additionally, we standardize all point clouds to 16,384 points to eliminate the effect of point cloud density. The generation time is also reported to reflect the efficiency of each method. Table~\ref{fig:rec_result} presents the visualization of our method.

\begin{table}[ht]
\caption{PERFORMANCE COMPARISON OF SINGLE-VIEW 3D RECONSTRUCTION}
\label{tab:performance_comparison}
\centering
\resizebox{\columnwidth}{!}{ 
\begin{tabular}{lccc}
    \toprule
    \textbf{Methods} & \textbf{CD $\downarrow$} & \textbf{FS $\uparrow$} & \textbf{Time (s) $\downarrow$} \\
    \midrule
    \textbf{Shape-E \cite{c45}} & 82.92 & 0.5533 & 48.26 \\
    \textbf{One-2-3-45 \cite{c46}} & 49.47 & 0.4927 & 56.40 \\
    \textbf{InstantMesh \cite{c47}} & \cellcolor{darkorange}30.38 & \cellcolor{darkorange}0.6446 & 20.14 \\
    \textbf{Direct3D \cite{c58}} & 37.75 & 0.5931 & \cellcolor{lightorange}17.93 \\
    \textbf{Triplane-Grasping} & \cellcolor{lightorange}31.85 & \cellcolor{lightorange}0.6188 & \cellcolor{darkorange}5.72 \\
    \bottomrule
\end{tabular}
}
\end{table}

As shown in Table~\ref{tab:performance_comparison}, our method, Triplane-Grasping, achieves a competitive trade-off between reconstruction accuracy and efficiency. We significantly outperforming all other methods in terms of generation time, which is over 2 times faster than Direct3D (17.93s) and more than 8 times faster than Shape-E (48.26s), while still maintaining the second-best Chamfer Distance (CD) and F-Score (FS), with only a marginal difference of 1.47 and 0.0258 compare to InstantMesh. However, our method requires only one-fourth of its generation time compare to InstantMesh, demonstrating significantly improved efficiency. This superior balance between speed and quality enables fast and reliable grasp decision making in real-world applications.

\subsection{6-DoF grasping decision evaluation}

To evaluate the generalization capability of our method for grasp prediction on unseen objects, we first conduct experiments on the GraspNet-1Billion dataset. Following common practice in prior works, we report grasping performance on three categories of objects: seen, similar, and novel, as defined by the official GraspNet-1Billion split. We also evaluate our method on the test set of OmniObject3D and report the grasp success rate in a simulated environment. The results are summarized in Table~\ref{tab:grasp_success_rate}.

It is worth noting that, in order to better isolate and analyze the generalization ability of our model, we evaluate grasping performance in a single-object tabletop setting. This allows for a clearer and more controlled analysis of object-level grasp prediction. Importantly, our model is trained solely on the OmniObject3D dataset, without any exposure to objects from GraspNet-1Billion during training. Therefore, although we adopt the dataset's original categories split for consistency in reporting, all test objects are effectively novel to our model. All reported average grasp precision is under a friction coefficient of 1.0. Figure~\ref{fig:vis} illustrates the qualitative results of our predicted grasps.

\begin{table}[ht]
\centering
\caption{Grasp success rate and inference time on GraspNet-1Billion and OmniObject3D.}
\label{tab:grasp_success_rate}
\renewcommand{\arraystretch}{1.1}
\setlength{\tabcolsep}{3pt}
\resizebox{\columnwidth}{!}{%
\begin{tabular}{c|ccc|c|c}
\toprule
\multirow{2}{*}{\textbf{Methods}} & \multicolumn{3}{c|}{\textbf{GraspNet-1Billion}} & \multirow{2}{*}{\textbf{OmniObject3D}} & \multirow{2}{*}{\textbf{Time (s)}} \\
\cmidrule{2-4}
& Seen & Similar & Novel & & \\
\midrule
GraspNet-Baseline~\cite{c59} & 35.72 & 31.98 & 17.82 & 10.38 & 22.51 \\
HGGD~\cite{c60} & \cellcolor{lightorange}63.67 & \cellcolor{lightorange}58.73 & \cellcolor{lightorange}24.86 & 27.96 & \cellcolor{lightorange}11.37 \\
GSNet~\cite{c61} & \cellcolor{darkorange}\textbf{65.01} & \cellcolor{darkorange}\textbf{59.10} & 23.82 & \cellcolor{lightorange}29.32 & 12.75 \\
\textbf{Ours} & 45.78 & 40.46 & \cellcolor{darkorange}\textbf{27.18} & \cellcolor{darkorange}\textbf{31.16} & \cellcolor{darkorange}\textbf{7.52} \\
\bottomrule
\end{tabular}
}
\end{table}

As shown in Table~\ref{tab:grasp_success_rate}, our method achieves competitive grasp success rates across all object categories, and particularly outperforms all baselines on the novel object subset of GraspNet-1Billion. This result underscores the strong generalization ability of our approach, especially in scenarios where no prior knowledge of the test objects is available. Notably, while other methods leverage depth information during training and inference, our method relies solely on RGB input, yet still achieves superior performance. This highlights the model’s strong capability in capturing object scale and spatial cues from RGB alone, making it particularly well-suited for tabletop grasping tasks. Furthermore, our method achieves the highest success rate on the OmniObject3D test set and demonstrates the fastest inference speed among all compared approaches, confirming both its effectiveness and practical efficiency.

\section{CONCLUSIONS}

We proposed Triplane Grasping, a fast grasp pose generation method which only need one single RGB-Only image. Our method achieves real-time grasp decision-making by introducing an efficient 3D reconstructs stage with the help of the hybrid Triplane-Gaussian representation along with a grasp generate model, providing robustness point cloud representation while ensuring the quality of grasping prediction. Our experiments showcases that the Triplane Grasping shows reliable and efficient grasping decision making performance when facing common tabletop objects. Future work includes extend this work to deal with more complex Cluttered scenes and real-world objects with a wider range of scale.







\end{document}